\documentclass{article}
\usepackage{spconf,amsmath,graphicx}

\usepackage{enumitem}
\setlist{nosep, leftmargin=14pt}
\usepackage[misc]{ifsym} 
\usepackage[skip=1pt]{caption}
\usepackage{mwe} 
\usepackage{array}

\title{Rapid Model Transfer for Medical Image Segmentation via \\ Iterative Human-in-the-loop Update: from Labelled Public to Unlabelled Clinical Datasets for Multi-organ Segmentation in CT}

\name{Wenao Ma$^{1}$, Shuang Zheng$^{2, 3}$, Lei Zhang$^{2, 3}$, Huimao Zhang$^{2, 3 \star}$\thanks{$\star$ \; \textit{Corresponding authors}}, Qi Dou$^{1 \star}$}
\address{$^{1}$ Dept. of Computer Science and Engineering, The Chinese University of Hong Kong, Hong Kong, China\\
$^{2}$ Department of Radiology, The First Hospital of Jilin University, Jilin, China \\
$^{3}$ Jilin Provincial Key Laboratory of Medical Imaging \& Big Data, Jilin, China}
\begin{document}
%
\maketitle
\begin{abstract}

\end{abstract}
Despite the remarkable success on medical image analysis with deep learning, it is still under exploration regarding how to rapidly transfer AI models from one dataset to another for clinical applications.
This paper presents a novel and generic human-in-the-loop scheme for efficiently transferring a segmentation model from a small-scale labelled dataset to a larger-scale unlabelled dataset for multi-organ segmentation in CT. To achieve this, we propose to use an igniter network which can learn from a small-scale labelled dataset and generate coarse annotations to start the process of human-machine interaction. Then, we use a sustainer network for our larger-scale dataset, and iteratively updated it on the new annotated data. Moreover, we propose a flexible labelling strategy for the annotator to reduce the initial annotation workload. The model performance and the time cost of annotation in each subject evaluated on our private dataset are reported and analysed. The results show that our scheme can not only improve the performance by 19.7\% on Dice, but also expedite the cost time of manuallabelling from 13.87 min to 1.51 min per CT volume during the model transfer, demonstrating the clinical usefulness with promising potentials.




\section{Introduction}

Despite current success of artificial intelligence technique for medical image diagnosis applications, some even promisingly embedded into the clinical practice, it is still very expensive, time-consuming and unstandardized if one wants transfer the AI tools from one dataset to another. Usually, the deep learning networks, as a data-driven approach, are prone to performance drop when being tested on images that have different distributions from the training dataset(s). To handle data discrepancies across scanners, protocols or demographics, solutions of transfer learning, domain adaptation and generalization have been studied, however, few investigations have emphasized clinician-AI interactions so far, and suffer from limitations on either efficiency or precision during model transfer.


\begin{figure}[t]
    \centering
	\includegraphics[width =9.0cm]{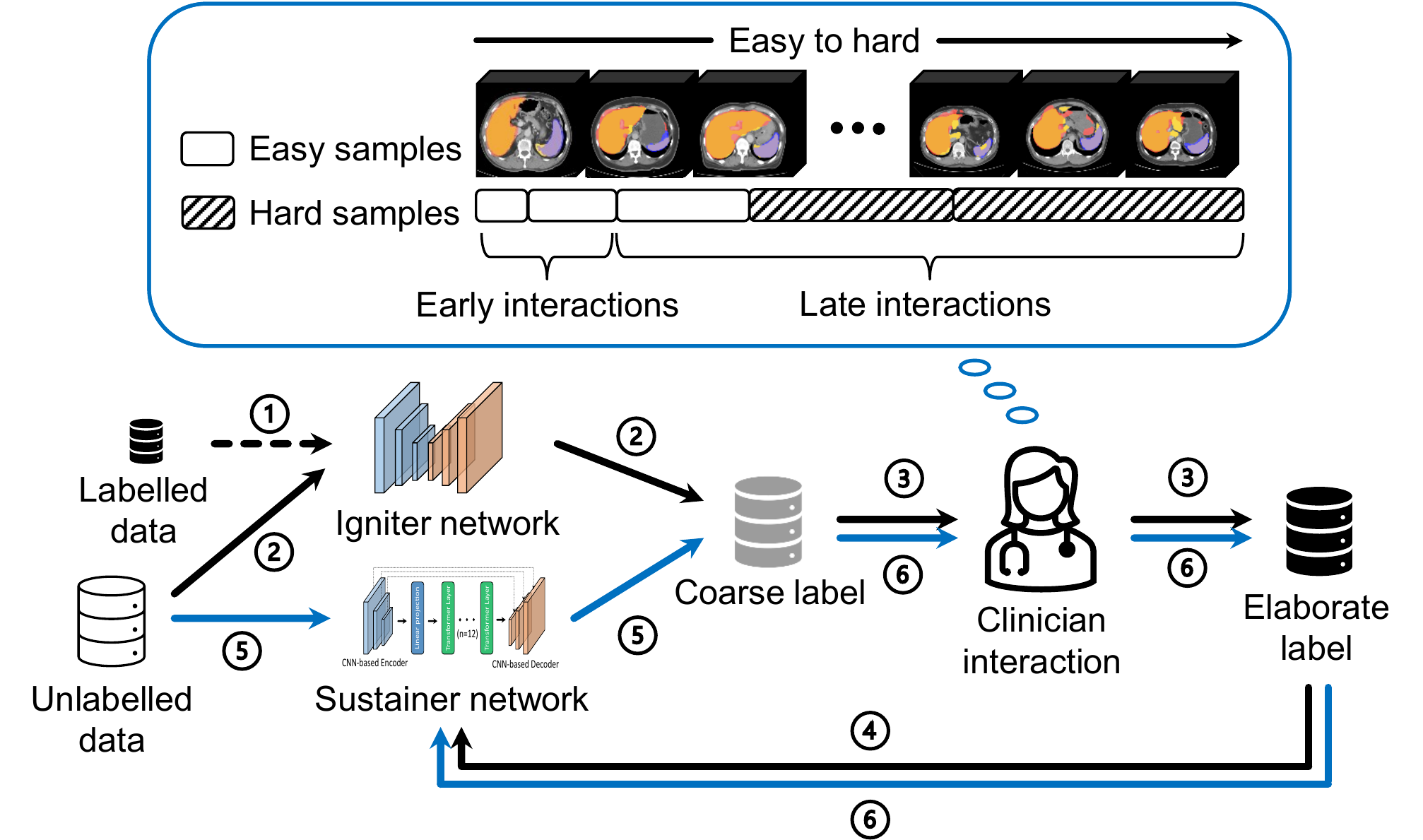}
	\caption{Overview of the proposed human-in-the-loop scheme: the coarse annotations on the unlabelled dataset are generated by the trained igniter network and are further refined by the clinician's efficient editions; then the sustainer network is iteratively trained on the new labelled data, executing an annotation loop between the clinician and AI model.} 
	\label{pipeline}
\vspace{-0.5cm}
\end{figure}


Recent favor of \emph{human-in-the-loop (HITL)} mechanisms is a new concept to address this challenge by incorporating expert knowledge to facilitate model update during optimization~\cite{budd2021survey}.
For examples, Amrehn \textit{et al.}~\cite{amrehn2017ui} propose to use an active user model which makes human iteratively add additional hints in a form of pictorial scribbles during training for segmentation models. Wang \textit{et al.}~\cite{wang2018deepigeos} present a novel framework that uses two networks, called P-Net and R-Net, i.e., the P-Net performs an initial segmentation while the R-Net takes the initial segmentation and the user interactions as the inputs and then provide a refined result. Jang and King \cite{jang2019interactive} propose an interactive segmentation algorithm which converts use-annotations into interaction maps to be the input of CNN and develop the backpropagating refinement scheme to correct the mislabelled pixels in nature images. The experiments also showed that their scheme can also be successfully applied to medical image segmentations. Liao \textit{et al.} \cite{liao2020iteratively} propose a dynamic process for successive interactions, which models interactive medical image segmentation as a Markov decision process. 
This dynamic process significantly outperforms other state-of-the-art approaches with fewer interactions.


To date, the methods mentioned above mainly focused on iteratively refining a segmentation model to achieve the result with a desired level of quality. These approaches assume that expert annotators can always generate labels when the model requires it. However, in clinical practice, we cannot overlook the availability of expert annotators as their time and cost are also extremely important. Lutnick \textit{et al.} \cite{lutnick2019integrated} has reported that the HITL strategy can significantly reduce the annotation effort, while Kuo \textit{et al.}~\cite{kuo2018cost} propose to jointly consider the informativeness and labelling time to pick samples to label. But, neither of them emphasized how to systematically minimize the time of manual labelling on a new dataset which is a key point for human-in-the-loop pipelines.

In this paper, we propose a novel and generic HITL framework to transfer a model from a small-scale labelled dataset to a larger-scale unlabelled dataset. Specifically, we develop an igniter network trained by a public available dataset, which can generate the coarse annotations on a new dataset and start the clinician-AI interaction. A sustainer is then adopted to execute the human-in-the-loop process, which can be iteratively updated by the new unlabelled dataset, executing an annotation loop between clinicians and AI models. More importantly, we design a labelling strategy to incorporate clinician editions for the model predictions while minimizing the time of manual labellings. In this study, we focus on evaluating the proposed framework in the task of CT liver and spleen segmentation. The results show that our framework can significantly improve the performance by 19.7\% on Dice with the help of clinician-AI interaction, while also expedite the cost time of manual labelling from 13.87 min to 1.51 min per CT volume during the model transfer. 
\label{sec:intro}

\section{Method}
Our problem setting is as follows: given a small-scale labelled seed set \textit{S} and a larger-scale unlabelled target set \textit{U}, rapidly generate the accurate pixel-wise segmentation labels of set \textit{U} via an iterative human-in-the-loop process, and finally train a good segmentation model for this set. An overview of our proposed framework is shown in Fig.~\ref{pipeline}. 

\subsection{Selection of igniter network and sustainer network}
\label{sec:format}
The first step of the proposed labelling scheme is to train an igniter network on a small-scale labelled seed set. As the igniter model is used to generate the coarse annotations on the unlabelled set \textit{U}, the selection of this igniter network can be quite flexible. The architecture of this igniter network should be adopted for the existing labelled data, with a high level of effectiveness and robustness. In this paper, we focus on multi-organ (i.e., liver and spleen) segmentation from CT images. In this regard, we adopt our recent multi-modal segmentation network~\cite{dou2020unpaired} which is trained with knowledge distillation on unpaired CT and MRI images. We chose this model as the igniter mainly for three reasons. First, the network~\cite{dou2020unpaired} can yield a superior performance on a small-scale dataset by wisely exploiting multi-modal information. Second, it is often the fact that people need to transfer an existing relevant model previously got from a public dataset to a clinical dataset coming up at hand. We just hereby mimic such a natural situation. Third, we show the flexibility of an igniter network, which not necessarily originates from an identical task to the targeting one.

After training the igniter network, the coarse annotations can be generated and further refined via the input of experts. A sustainer network is then trained on these new labelled data, executing an annotation loop between clinicians and AI models. Next, the unlabelled data are annotated by the sustainer network iteratively and further refined by the human-in-the-loop interventions. There are existing two main approaches used to train a model on new annotations, including retraining the model on all available data, and fine-tuning the model. 
We go for the former approach, which is computationally more expensive but can make the full use of the clinical cohort via sufficient training. For network architecture, we use the state-of-the-art TransUNet~\cite{chen2021transunet} which is designed for multi-organ segmentation using transformers. 
Note that, in our HITL scheme, the model architectures of the igniter and sustainer are not necessarily to be the same. We hereby adopt the latest transformer-based model for the sake of its high performance, based on sufficient training with our collected relatively large clinical dataset.

\subsection{Iterative labelling strategy}
\label{sec:format}
To design an iterative labelling strategy with the aim of minimizing the time of manual labelling for the whole unlabelled set \textit{U}, it is important to properly order which samples to annotate first for the interactions. Different from previous methods which conduct the selection based on calculation-based measurements, we respect more for human factors. Based on discussions with radiologists, we make it flexible for them to make the setting of interactions, with intuitive guidelines from following considerations.

1) \textit{Update the model frequently in the early interactions.} In the beginning of model updates, the time cost of manual labelling of each sample is pretty high, while in turn, each sample also has the potential to bring larger improvement to the segmentation model on the new dataset. These factors inspire us to update the model more frequently in the early rounds of interactions. In other words, we allow the experts to annotate a small number of samples while the remaining unlabelled data can be annotated later after the model is upgraded, which expedite the manual labelling process overall. 

2) \textit{Choose the samples which are easy to annotate in the early interactions.} Intuitively, the samples with poor annotation quality have more potentials to improve the model performance. In addition, in the field of active learning, the model usually choose the samples being rich in informativeness to annotate. However, the samples with these properties are usually time-consuming to be refined manually, which leads to the efficiency dilemma. We therefore have performed experiments to explore the best strategy to save the labelling time of experts. The results will be shown in the experiment section, which indicates that, surprisingly, the most effective way is to select easy samples to annotate in the early interactions. Given the limited time in the early annotation iteration, the experts can annotate more samples if they select the samples which are well segmented and easier to be refined. Meanwhile, as the model improved iteratively, the process of labelling hard samples can be expedited in later annotation iterations. Therefore, choosing the easy samples to annotate to improve the model performance efficiently in the early interactions is encouraged in our proposed scheme.

\section{Experimental results}
\label{sec:pagestyle}

For experimental validation, we perform liver and spleen segmentation in CT images, which is an important prerequisite for clinical downstream applications such as automated liver fibrosis staging \cite{choi2018development, son2020assessment}. To demonstrate the effectiveness of the proposed scheme, we evaluate the model on our private dataset which was not initially annotated.


\subsection{Datasets and implementation details}
\label{sec:format}



\textbf{Labelled set:} We train the igniter network on public datasets, i.e., a CT dataset with 30 patients and a MRI dataset with 20 patients, which both come from ISBI 2019 CHAOS Challenge~\cite{kavur2021chaos}. The original images are cropped at the areas of multi-organs, and are both resampled into around 1.5 $\times$ 1.5 $\times$ 8.0 $mm^{3}$, with a size of 256 $\times$ 256 in transverse plane. For the preprocessing, the intensity values of raw images are truncated to the range of [-200, 250], and are further normalized to zero-mean and unit variance.

\textbf{Unlabelled set:} We collected an in-house dataset of Contrast Agent-enhanced Portal Venous Phase abdominal CT images from The First Hospital of Jilin University. Our dataset consists of 551 patients, who were diagnosed with liver fibrosis, and underwent liver biopsy or liver resection during 2016 and 2021. The inclusion criteria are: the age being 18 years or older, no previous liver surgery, availability of high-quality CT images, no tumor that changes the hepatic morphological. The CT images have slice thickness of 5 $mm$ with in plane resolution of 512 $\times$ 512. The preprocessing used in this in-house set is the same with the preprocessing of the labelled set. The human-in-the-loop annotations were conducted by a radiologist with over 10 years of experience.

\textbf{Implementation details:} The igniter network is trained with Adam optimizer, while the batch size is set as 4 and the initial learning rate is set as $0.0001$. The learning rate is decayed by $5\%$ in every 500 iterations and the total number of training iterations is $30k$. For the sustainer network, the patch size of transformer is set as 16 and batch size is set as 12. The model is trained with SGD optimizer, the learning rate is initialized as $0.01$ and the training epoch is 20 for each model update. The other implementation details are the same as described in \cite{dou2020unpaired} and \cite{chen2021transunet}. 

\begin{figure}[!t]
\vspace{-0.1cm}
    \centering
	\includegraphics[width=8.5cm]{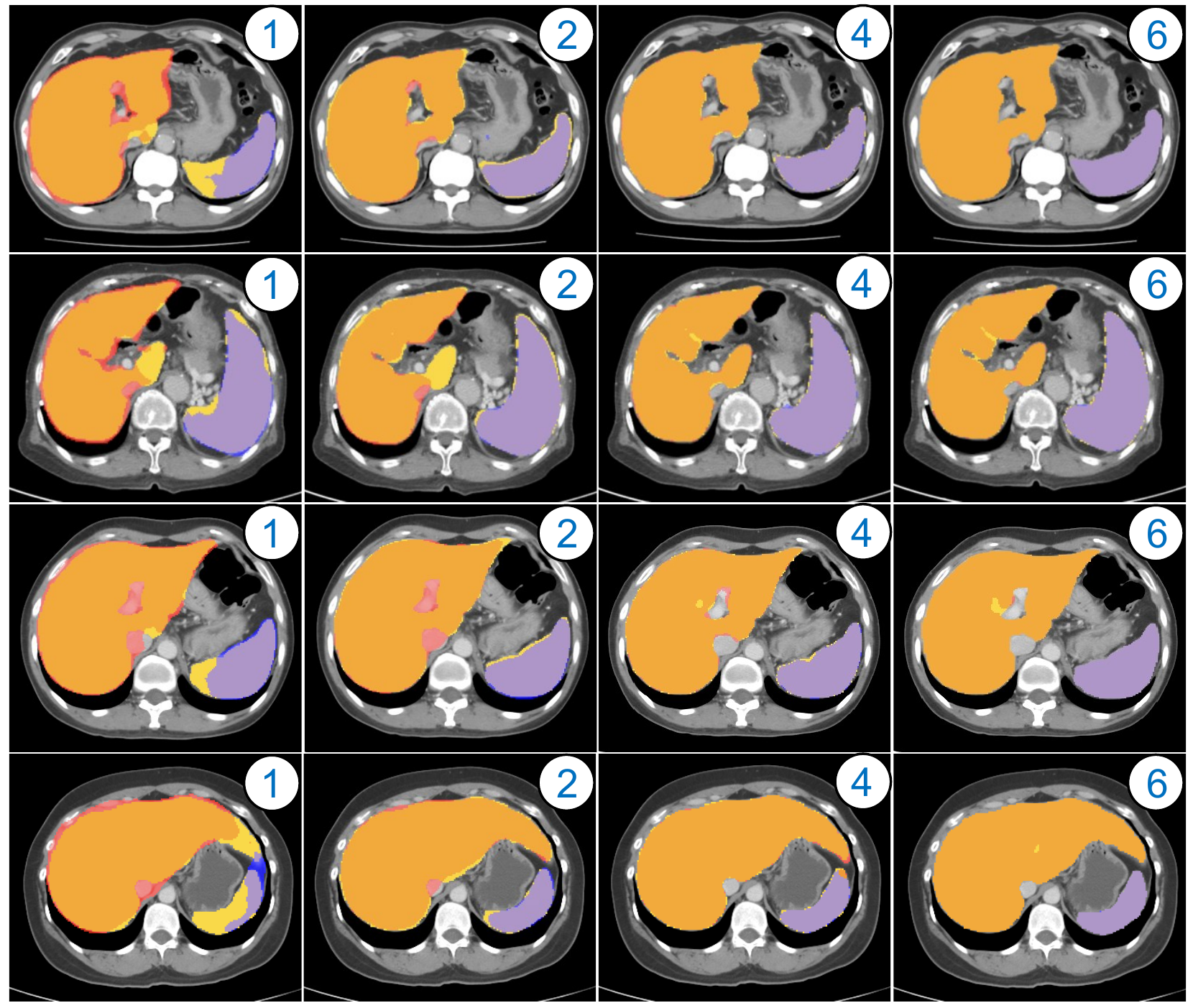}
	\caption{Visualization of model segmentation results after the 1st, 2nd, 4th, and 6th rounds of update. The yellow region indicates the ground truth of liver and spleen. The red and blue regions indicate the model predictions of liver and spleen respectively. The orange and purple regions indicate the overlap of predicted regions with the ground truth.} 
	\label{visualization}
 \vspace{-0.6cm}
\end{figure}


\begin{figure*}[t]
    \centering
    
	\includegraphics[width =18.0cm]{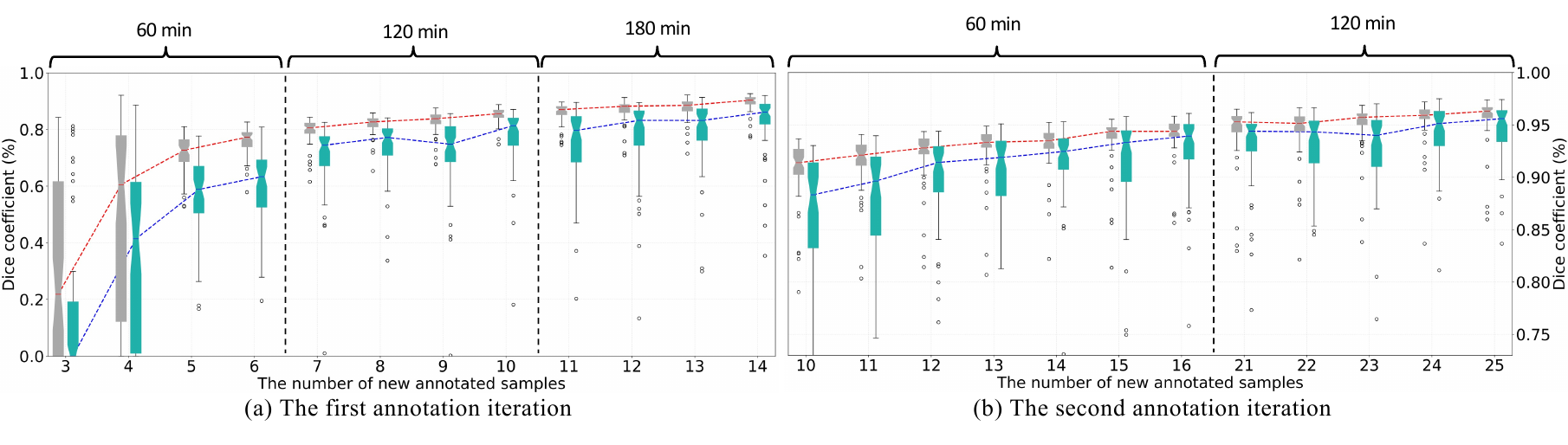}
	\caption{Labelling strategy analysis in the first and second annotation iteration. The grey boxes show the liver segmentation results while the green boxes show the spleen segmentation results. The medians of grey boxes and green boxes are connected with red dotted line and blue dotted line, respectively. For each fixed manual labelling time, the number of new annotated samples is increased successively to compare the model performance of selecting easy samples and that of selecting hard samples.} 
	\label{annotation_analysis}
\vspace{-0.4cm}
\end{figure*}

\begin{figure}[t]
    \centering
	\includegraphics[width=0.48\textwidth]{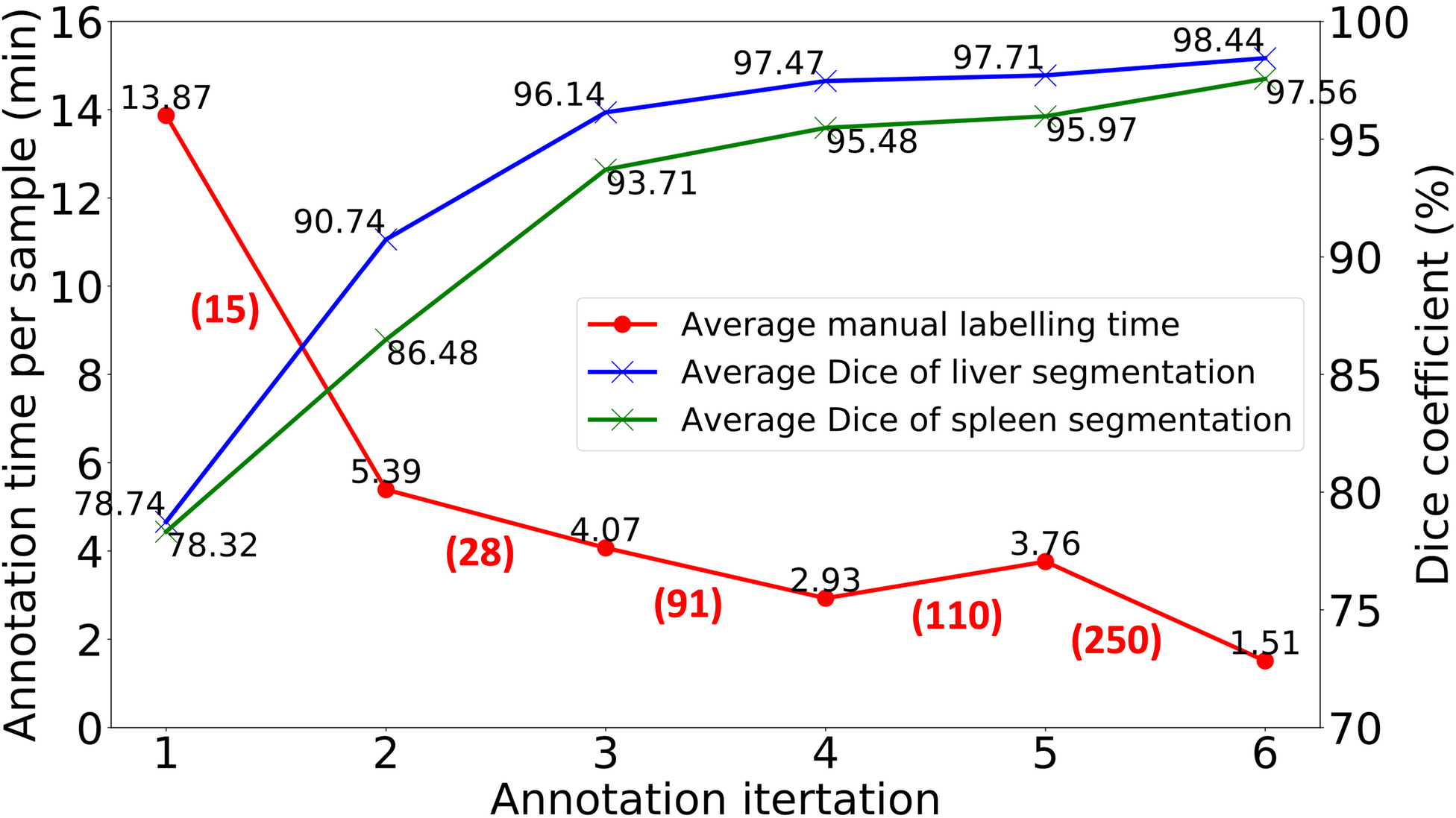}
	\caption{Evaluation of Dice for segmentation performance for human-in-the-loop model updates, with the annotation time reported (red line). The numbers in brackets are the number of manually labelled cases for each iteration.} 
	\label{quantification}
\end{figure}

\subsection{Evaluation on segmentation performance for human-in-the-loop model updates}
\label{sec:format}
\vspace{-0.3cm}

We extensively evaluate the effectiveness of the proposed HITL framework on our clinical dataset. We first train the igniter network on the labelled set and then iteratively train the sustainer network. For each training round, an experienced radiologist manually edit the model predictions, for providing the ground truth to further update the sustainer network. It is worth to mention that the number of samples to annotate for each interaction round are adopted for expert's practical schedule. The time cost of manual labelling of all the samples and their averages are recorded.

We evaluate the segmentation performance of the model by the Dice Coefficient at each annotation iteration, using the 57 samples refined at the last annotation iteration to be the test set. As can be seen in Fig.~\ref{visualization} and Fig.~\ref{quantification}, the model performance increases rapidly in the early interactions, which also benefits the decrease of the average manual labelling time. For example, after the first iteration in which the expert only edited labels for 15 cases (taking around 14 mins each), the segmentation Dice jumped from 78.74\% to 90.74\% (increasing 12.00\%) for liver, and from 78.32\% to 86.48\% (increasing 8.16\%) for spleen.
In the following interactions, the model performance is kept gradually increasing, and more importantly, the manual annotation time is stable at a reasonable level (less than 5 mins).
Though the remaining samples in the later interactions are relatively harder to be refined (according to our designed protocol in Sec. 2.2), the average manual labelling time still does not fluctuated wildly. These observations demonstrate that updating the model frequently in the early interactions, even with a small amount of data, can be an efficient way to save the manual annotation time of the expert. 

\subsection{Ablation study on labelling strategy}
\label{sec:format}

We further perform experiments to analyze the strategy of sample selection in the early interactions. As we have already recorded the manual time cost of each sample, we can compare the models performance which trained by different selections of samples. In this part, we fix the total manual labelling time of each interaction, which can be 60, 120 or 180 minutes. Given the fixed manual labelling time, we select different samples with different time cost of labelling, so that the summation of labelling time of these samples can equal to the fixed time, i.e., 60, 120 or 180 minutes. In other words, as the number of new annotated samples increases, it indicates we use more easy samples to update the model during the AI-clinician interaction. The results of the first interaction and second interaction can be seen in Fig.~\ref{annotation_analysis}. For example, in the first annotation iteration, when the expert only has 60 minutes to edit the labels, if he or she selects the hardest samples to annotate, only 3 samples with elaborate labels can be provided to update the model. On the contrary, if the expert selects the easiest samples to edit, he or she can generate 6 new samples to update the model, which makes the Dice Segmentation on test set increase sharply and become more stable compared to the former selection. These results show that the best strategy to select the samples in the early interaction is to choose the easy samples and try to annotate as more samples as possible when given the limited time, which further verifies the soundness of our designed labelling strategy.

\vspace{-0.4cm}

\section{Conclusion and Future Work}
\label{sec:typestyle}
In this paper, we propose a simple yet effective scheme to leverage human-in-the-loop mechanism for rapidly transferring an AI model trained on a labelled small-scale dataset to an unlabelled larger-scale dataset for a specific clinical application.
The experiments show that our proposed method can not only improve the model performance on the targeting task, but also save the time of manual labelling, demonstrating promising value in the real-world clinical scenarios. Our future work will continue to use this HITL scheme to label more of our collected data, and further apply the automated segmentation results for the downstream clinical applications such as intelligent liver fibrosis staging.



\section{COMPLIANCE WITH ETHICAL STANDARDS}
This research study was conducted retrospectively using human subject data from The First Hospital of Jilin University with ethics approval.

\section{ACKNOWLEDGMENTS}
This work was supported by the CUHK Shun Hing Institute of Advanced Engineering (project  MMT-p5-20), CUHK Direct Research Grant, Shenzhen-HongKong Collaborative Development Zone, and China International Medical Foundation, Imaging Research, SKY (project Z-2014-07-2003-03).







\end{document}